\documentclass[a4paper]{cas-sc}   
\usepackage[numbers]{natbib}
\usepackage[labelfont=bf]{caption}
\usepackage{amsmath,amsfonts,amssymb,amsthm,booktabs,graphicx,url} 
\usepackage{multirow}
\usepackage{mathrsfs}
\usepackage{algorithm, algorithmic}
\usepackage{microtype}
\usepackage{enumitem}
\usepackage{makecell}
\theoremstyle{plain}

\theoremstyle{definition}

\theoremstyle{remark}
\newtheorem{rem}{Remark} 

\usepackage{float}

\begin{document}

\shorttitle{FDRMFL: Multimodal Federated Feature Extraction Model}
\shortauthors{Wu H.}

\title[mode = title]{FDRMFL: Multimodal Federated Feature Extraction Model Based on Information Maximization and Contrastive Learning}

\author[1]{Haozhe Wu}
\address [1]{School of Computing and Data Science, The University of Hong Kong, Hong Kong SAR, China} 
\cortext[mycorrespondingauthor]{Corresponding author. Email address: \texttt{haozhe\_wu@connect.hku.hk}}

\begin{abstract}
We propose FDRMFL, a task-driven multimodal feature extraction framework for federated regression under non-IID data distributions.
Extracting predictive features from high-dimensional multimodal inputs is particularly challenging in this setting: data cannot leave each client, local samples are scarce and heterogeneously distributed, and unsupervised dimensionality reduction discards task-relevant information while federated training introduces representation drift across communication rounds.
FDRMFL addresses these challenges through a unified four-term local objective: MSE prediction loss, a correlation-based mutual information surrogate that preserves dependence between the fused representation and the continuous target, a symmetric KL penalty that aligns cross-modal latent distributions before fusion, and an InfoNCE-style contrastive loss that anchors local representations to the global consensus.
Experiments on three synthetic and two real-world near-infrared spectroscopy datasets under non-IID federated partitions, with comprehensive ablation and sensitivity analyses, demonstrate that each component contributes to the framework's effectiveness.
FDRMFL reduces mean MSE by 33.8\% relative to the best traditional baseline (PCA) and by 43.0\% relative to VAE in simulation, and attains the lowest overall mean MSE among six federated algorithms including FedAvg, FedProx, MOON, SCAFFOLD, and FedBN.
\end{abstract}


\begin{highlights}
\item A task-driven multimodal federated feature extraction framework (FDRMFL)
      that jointly optimizes prediction accuracy and representation quality
      under non-IID data across clients.
\item A multi-objective local training procedure combining MSE prediction loss,
      correlation-based mutual information retention, symmetric KL cross-modal
      alignment, and contrastive representation consistency.
\item Comprehensive evaluation on three synthetic and two real-world
      near-infrared spectroscopy datasets demonstrating consistent MSE
      reduction over PCA, TSVD, RP, and VAE baselines and the lowest
      overall mean MSE among six federated algorithms.
\item Ablation analysis confirming that each regularization component
      contributes to overall performance and cross-client stability.
\end{highlights}

\let\printorcid\relax
\maketitle

\section{Introduction}

Multimodal data---images, spectra, time series, text---arise naturally across scientific and engineering domains where a single modality cannot capture the full complexity of the underlying phenomenon \cite{dey2022computational,muscat2024functional}. Representative applications span evolutionary biology \cite{billera2001geometry}, developmental biology \cite{schiebinger2019optimal}, neuroimaging \cite{arsigny2007geometric,dryden2009non,yuan2012local}, network science \cite{dubey2022modeling}, medical imaging \cite{marron2021object}, and social information analysis \cite{daradkeh2022tools}; see \cite{Baltrusaitis2019} for a comprehensive survey. Because different modalities often reside in distinct topological or functional spaces whose inherent correlations violate Euclidean-distance assumptions, extracting compact and predictive representations from multimodal inputs is a prerequisite for accurate downstream analysis---yet substantially harder than in the single-modality case \cite{Baltrusaitis2019}.

The extraction problem becomes substantially harder when data are distributed across institutions that cannot share raw observations. Federated learning \cite{McMahan2017,Kairouz2021,Wang2021FieldGuide} enables collaborative model training under such privacy constraints, but non-IID local distributions introduce representation drift, cross-modal misalignment, and the curse of dimensionality---especially in high-dimensional regression with limited samples \cite{Li2020FedProx,Li2021FedBN,Xiong2022FedMM,Kirkpatrick2017,Alotaibi2024NonIID}.

Existing feature extraction approaches exhibit significant limitations in this setting. \emph{Unsupervised linear reductions}---Principal Component Analysis (PCA), Truncated SVD (TSVD), and Random Projection (RP)---optimize variance preservation or distance retention while ignoring downstream labels entirely: PCA discards low-variance directions that may carry strong predictive signals \cite{Bair2006}; RP randomly projects away task-relevant features \cite{Bingham2001}; and TSVD performs low-rank compression without modal alignment \cite{Halko2011}. These methods further assume centralized, identically distributed data; under cross-client non-IID distributions, local feature subspaces shift substantially, damaging model consistency \cite{Karimireddy2020,Kirkpatrick2017}. Even Variational Autoencoders (VAE), which learn nonlinear representations, optimize a reconstruction rather than a prediction objective, and aligning their latent spaces across federated clients remains nontrivial \cite{higgins2017betavae}. On the federated optimization side, FedAvg \cite{McMahan2017}, FedProx \cite{Li2020FedProx}, SCAFFOLD \cite{Karimireddy2020}, MOON \cite{Li2021MOON}, and FedBN \cite{Li2021FedBN} address client drift through proximal penalties, variance reduction, or contrastive consistency, yet none provides built-in mechanisms for cross-modal alignment or task-driven feature retention in multimodal regression \cite{Feng2023FedMultimodal}. Recent work on multimodal representation alignment \cite{MERL2025IBMMultiModal} and deep sufficient modality learning \cite{gao2025deepsum} has advanced the centralized setting, yet no prior method jointly addresses task-driven dimensionality reduction for continuous targets, cross-modal distributional alignment, and cross-round representation consistency within a single federated framework.

We propose FDRMFL (Federated Dimensionality-Reducing Multimodal Feature extraction and Learning), a federated multimodal feature extraction framework. FDRMFL trains modality-specific encoders, a cross-attention fusion network, and a regression head with a unified four-term local objective that combines MSE prediction loss with three regularizers targeting task-relevant dependence, cross-modal alignment, and cross-round consistency (Section~\ref{section2}; Figure~\ref{fig:overall_arch}).

The core technical contributions are threefold. We introduce a correlation-based mutual information surrogate $\mathcal{R}_{\mathrm{mi}}$ that maximizes dependence between the fused representation and the continuous target; motivated by the Gaussian MI identity $I(X;Y)=-\tfrac{1}{2}\log(1-\rho^2)$~\cite{Cover2006}, maximizing $|\hat\rho|$ provides a low-variance objective suited to the small-sample federated regime. We design a federated contrastive loss $\mathcal{R}_{\mathrm{fcl}}$, inspired by MOON~\cite{Li2021MOON} but operating in the fused multimodal space~\cite{Oord2018CPC} with a temporal history buffer of negatives, to anchor local representations to the global consensus. We further impose a symmetric KL penalty $\mathcal{R}_{\mathrm{kl}}$ that closes the modality gap~\cite{Liang2022ModalityGap} by aligning latent distributions before fusion. Together, these components constitute the first unified framework for task-driven dimensionality reduction, cross-modal alignment, and representation stability in federated multimodal regression. Experiments on three synthetic and two real NIR spectroscopy datasets under non-IID partitions, with ablation and sensitivity analyses, show that FDRMFL reduces mean MSE by $33.8\%$ relative to the best traditional baseline (PCA) and by $43.0\%$ relative to VAE, and attains the lowest overall mean MSE among six federated algorithms.

The remainder of this paper is organized as follows. Section~\ref{section2} presents the FDRMFL formulation and its parameter estimation procedure. Section~\ref{section3} reports simulation and real-data experiments. Section~\ref{section4} concludes with a summary and directions for future work.

\begin{figure}[pos=H] 
  \centering
  \includegraphics[width=0.7\textwidth]{1.png}
  \caption{Overall architecture for federated multimodal learning.}
  \label{fig:overall_arch}
\end{figure}

\section{Methodology}\label{section2}
\subsection{Multimodal federated feature extraction}
Consider a federated learning setting with $K$ clients, where client~$i$ has local data distribution $P_i$ and sample weight $\pi_i$ ($\sum_{i=1}^K \pi_i = 1$), so that the global distribution is $P = \sum_{i=1}^K \pi_i P_i$. Each client holds multimodal inputs $X_i = (X_{i1}, X_{i2}, \ldots, X_{iM})$, where $X_{im}$ denotes the $m$-th modality (e.g., text, image, time series). Modality-specific neural encoders map each input to a $d$-dimensional feature vector---for instance, a Transformer for text ($z_{i,\text{text}} = h_{\text{text}}(X_{i,\text{text}})$), a CNN for images ($z_{i,\text{img}} = h_{\text{img}}(X_{i,\text{img}})$), or an LSTM for sequential data ($z_{i,\text{seq}} = h_{\text{seq}}(X_{i,\text{seq}})$).

The fusion function $g$ maps the collection of modality-specific
features $\{z_{i1}, z_{i2}, \ldots, z_{iM}\}$ to a unified
representation $Z_i \in \mathbb{R}^d$. In our implementation,
$g$ consists of three stages. Bidirectional multi-head cross-attention~\cite{Vaswani2017} is first applied between each pair of modality features, enabling each modality to attend to and incorporate information from the others. A learned attention mechanism then computes sample-dependent importance weights $\alpha_m$ for each modality via a softmax gate, producing weighted features $\hat{z}_{im} = \alpha_m \tilde{z}_{im}$. Finally, the weighted features are concatenated and passed through a batch-normalized MLP, $Z_i = \mathrm{MLP}_g([\hat{z}_{i1}; \ldots; \hat{z}_{iM}])$, followed by a residual self-attention refinement layer.
All parameters of $g$ are included in the global parameter set $w$
and participate in federated aggregation.

The prediction head $f$ maps the fused representation to a scalar output:

\begin{equation}
y_i = f(Z_i) + \epsilon_i
\end{equation}

where $\epsilon_i$ is a zero-mean noise term ($\mathbb{E}[\epsilon_i | Z_i] = 0$, $\text{Var}(\epsilon_i | Z_i) = \sigma_{\epsilon,i}^2$). The global objective minimizes a weighted sum of four terms---prediction loss, mutual information regularization, modality alignment, and federated contrastive regularization---defined as follows. The \emph{prediction loss} measures the deviation between model output and target:
\begin{equation}
\mathcal{L}_{\text{pred}} = \mathbb{E}_{(X_i, y_i) \sim P} \left[ (y_i - f(Z_i))^2 \right]
\end{equation}
Since $y_i = f(Z_i) + \epsilon_i$, this equation can be expanded as
\begin{equation}
\mathbb{E}[\epsilon_i^2] = \sum_{i=1}^K \pi_i \left( \sigma_{\epsilon,i}^2 + \mathbb{E}[(f(Z_i) - \mathbb{E}[y_i | Z_i])^2 | P_i] \right),
\end{equation}
which decomposes into weighted client noise variance plus conditional bias, ensuring basic prediction capability.

The \emph{mutual information regularization} maximizes $I(Z_i; y_i)$ to ensure that the latent representation retains predictive information about the target:
\begin{equation}
I(Z_i; y_i) = \mathbb{E}_{P(Z_i, y_i)} \log \frac{P(Z_i, y_i)}{P(Z_i) P(y_i)}
\end{equation}
Equivalently $I(Z_i; y_i) = H(y_i) - H(y_i | Z_i)$, so the regularization term is
\begin{equation}
\mathcal{R}_{\text{mi}} = -I(Z_i; y_i)
\end{equation}
Equivalently, the mutual information can be written as
\begin{equation}
\mathbb{E}_{P(y_i)} \mathbb{E}_{P(Z_i | y_i)} \log \frac{P(Z_i | y_i)}{P(Z_i)}
\end{equation}
The divergence between $P(Z_i | y_i)$ and $P(Z_i)$ grows as $Z_i$ captures more information about $y_i$.

\begin{rem}[Practical estimation of mutual information]\label{rem:mi}
In practice, we adopt a \emph{dependence maximization surrogate} that is
monotonically related to mutual information under mild regularity conditions.
The specific empirical form is detailed in Section~\ref{sec:estimation}.
\end{rem}

The \emph{cross-modal alignment regularization} minimizes distributional differences between modality features to ensure consistent fusion. For modalities $m$ and $n$ of client~$i$, the symmetric KL divergence measures the distance between $p(z_{im})$ and $p(z_{in})$:
\begin{equation}
\begin{split}
\mathrm{symKL}\big(p(z_{im}), p(z_{in})\big)
&= \frac{1}{2} \Big(
    \mathrm{KL}\big(p(z_{im}) \Vert p(z_{in})\big) \\
&\quad
    + \mathrm{KL}\big(p(z_{in}) \Vert p(z_{im})\big)
  \Big)
\end{split}
\end{equation}

where the KL divergence is defined as:
\begin{equation}
\text{KL}(p \| q) = \mathbb{E}_p \log \frac{p(z)}{q(z)}
\end{equation}
Averaging over all modality pairs under the global distribution:
{\small
\begin{equation}
\mathcal{R}_{\mathrm{kl}} = \mathbb{E}_{(X_i, y_i) \sim P} \left[
  \frac{1}{\binom{M}{2}}
  \sum_{1 \le m < n \le M}
  \mathrm{symKL}\big(p(z_{im}), p(z_{in})\big)
\right]
\end{equation}
}

To obtain a tractable form, we adopt a Gaussian working assumption
for the modality feature distributions. Specifically, we model each
modality feature as $p(z_{im}) \approx \mathcal{N}(\mu_{im}, \sigma^2 I)$
with a shared, fixed variance parameter $\sigma^2 > 0$ across all
modalities. Under this assumption, the symmetric KL divergence
simplifies to a scaled squared Euclidean distance between the modality
means:
\begin{equation}
\mathrm{symKL}\big(p(z_{im}), p(z_{in})\big)
= \frac{1}{2\sigma^2}\|\mu_{im} - \mu_{in}\|^2
\end{equation}

\begin{rem}[Practical interpretation]\label{rem:kl}
The homoscedastic Gaussian assumption is adopted as a \emph{working
simplification} rather than a distributional claim. The Gaussian
maximizes entropy among all distributions with a given mean and
covariance~\cite{Cover2006}, making it the least informative
(most conservative) distributional assumption. The fixed variance
$\sigma^2$ is absorbed into the regularization coefficient
$\lambda_2$, effectively yielding a mean-matching regularization:
\begin{equation}
\mathcal{R}_{\mathrm{kl}} \propto
\mathbb{E}\!\left[\frac{1}{\binom{M}{2}}
\sum_{1 \le m < n \le M}
\|\mu_{im} - \mu_{in}\|^2\right]
\end{equation}
In Section~\ref{sec:sensitivity}, we provide a sensitivity analysis
varying $\lambda_2$ to demonstrate stable performance.
\end{rem}

The \emph{federated contrastive regularization} constrains local representations to remain close to the global consensus via an InfoNCE-style loss. Let $Z_i$ be the current local representation, $Z_{i,\text{prev}}^g$ the global model's representation from the previous round (positive sample), and $\{Z_{i,\text{hist}}^g\}$ representations from earlier rounds (negative samples):
\begin{equation}
\mathcal{R}_{\mathrm{fcl}}
= \mathbb{E}_{(X_i, y_i) \sim P} \! \left[
  -\log \frac{
    \exp\bigl( \mathrm{sim}(Z_i, Z_{i,\mathrm{prev}}^g) / \tau \bigr)
  }{
    \exp\bigl( \mathrm{sim}(Z_i, Z_{i,\mathrm{prev}}^g) / \tau \bigr)
    + \smash{\sum_{\substack{z \in \{Z_{i,\mathrm{hist}}^g\}}}}
      \exp\bigl( \mathrm{sim}(Z_i, z) / \tau \bigr)
  }
\right]
\end{equation}

where $\text{sim}(a, b) = a^\top b / (\|a\| \|b\|)$ is cosine similarity and $\tau > 0$ is the temperature. This loss maximizes similarity with the positive (recent global) representation and minimizes similarity with negatives (earlier global states), anchoring local updates to the global consensus.

\begin{rem}[Mechanism for drift mitigation]\label{rem:fcl}
The contrastive loss $\mathcal{R}_{\mathrm{fcl}}$ acts as a soft
regularizer that penalizes large deviations of the local
representation $Z_i$ from the previous global representation
$Z_{i,\mathrm{prev}}^g$. We treat
$\mathcal{R}_{\mathrm{fcl}}$ as an empirically effective mechanism
for representation stability, validated through ablation experiments
(Section~\ref{sec:ablation}).

The historical negative sample set $\mathcal{H}_i$ stores up to
$H$ representations from earlier global rounds (we use $H=5$ in
all experiments), with oldest entries discarded when the buffer is
full. This FIFO strategy ensures that negative samples reflect
recent but distinct model states. The additional storage
per client is $H \times b \times d$ floating-point values.
\end{rem}

\paragraph{Relationship to MOON.}
The federated contrastive regularizer shares its InfoNCE-style
formulation with MOON~\cite{Li2021MOON}. However, FDRMFL differs
in three respects.
\emph{First}, $\mathcal{R}_{\mathrm{fcl}}$ operates on the
\emph{fused multimodal representation} $Z = g(z_1, \ldots, z_M)$,
stabilizing not only individual encoder features but also the
learned cross-modal alignment.
\emph{Second}, MOON forms a positive pair with the \emph{current}
global representation and uses the \emph{previous} global as the
sole negative; FDRMFL pairs with the \emph{most recent} global
as positive and draws negatives from a temporal history buffer,
providing richer contrastive signal.
\emph{Third}, MOON targets single-modality classification with
only a contrastive term; FDRMFL targets multimodal regression and
adds $\mathcal{R}_{\mathrm{mi}}$ and $\mathcal{R}_{\mathrm{kl}}$
for challenges specific to multimodal regression.

Table~\ref{tab:method_comparison} provides a structured comparison
across the design dimensions highlighted by related work.

\begin{table}[H]
\centering
\caption{Structured comparison of federated methods.
``---'' = not addressed;
communication overhead is relative to \textsc{FedAvg}.}
\label{tab:method_comparison}
\resizebox{\columnwidth}{!}{%
\begin{tabular}{@{}lcccccc@{}}
\toprule
& \textsc{FedAvg} & \textsc{FedProx} & \textsc{SCAFFOLD}
& \textsc{MOON} & \textsc{FedBN} & \textbf{FDRMFL} \\
\midrule
Local objective
  & $\mathcal{L}_{\text{task}}$
  & $\mathcal{L}_{\text{task}}{+}\mu\|w{-}w^g\|^2$
  & \makecell{$\mathcal{L}_{\text{task}}$\\(gradient correction)}
  & $\mathcal{L}_{\text{task}}{+}\mathcal{L}_{\text{con}}$
  & $\mathcal{L}_{\text{task}}$
  & \makecell{Eq.~(13): MSE prediction loss\\+ MI regularization\\+ KL alignment\\+ FCL regularization} \\
Drift handling
  & ---
  & \makecell{Parameter-space\\proximal penalty}
  & \makecell{Gradient-variance\\reduction}
  & \makecell{Representation-space\\contrastive loss}
  & \makecell{Local batch\\normalization}
  & \textbf{\makecell{Fused multimodal\\representation contrastive\\regularization}} \\
Communication overhead
  & $1\times$
  & $1\times$
  & $2\times$
  & $1\times$
  & ${<}1\times$
  & $1\times$ \\
Multimodal fusion
  & ---
  & ---
  & ---
  & ---
  & ---
  & \textbf{\makecell{Cross-attention\\+ dynamic weighting}} \\
Supervision
  & Any
  & Any
  & Any
  & Classification
  & Any
  & \textbf{Regression} \\
Regression-specific design
  & ---
  & ---
  & ---
  & ---
  & ---
  & \textbf{\makecell{Correlation-based MI surrogate\\for continuous $y$}} \\
\bottomrule
\end{tabular}%
}
\end{table}

The global objective combines all four terms:
\begin{equation}
\mathcal{L}_{\text{total}} = \mathcal{L}_{\text{pred}} + \lambda_1 \mathcal{R}_{\text{mi}} + \lambda_2 \mathcal{R}_{\text{kl}} + \lambda_3 \mathcal{R}_{\text{fcl}}
\end{equation}
where $\lambda_1 \equiv \lambda_{\mathrm{mi}}$, $\lambda_2 \equiv \lambda_{\mathrm{kl}}$, $\lambda_3 \equiv \lambda_{\mathrm{fcl}} > 0$ are regularization coefficients (the subscripted notation is used hereafter).

\subsection{Parameter estimation}\label{sec:estimation}

In practice, population expectations are replaced by mini-batch estimates. For client~$i$ with batch $\mathcal{B}_i = \{(X_{ij}, y_{ij})\}_{j=1}^b$, the empirical prediction loss is:
\begin{equation}
\mathcal{L}_{\text{pred},i}^{\text{emp}} = \frac{1}{b} \sum_{j=1}^b (y_{ij} - f(Z_{ij}))^2
\end{equation}
where $Z_{ij} = g(z_{ij1}, z_{ij2}, \cdots, z_{ijM})$ is the fused representation of the $j$-th sample for the $i$-th client, and $z_{ijm} = h_m(X_{ijm})$ is its $m$-th modal feature.

For the mutual information term, we adopt a correlation-based dependence
surrogate. Let $f_\phi: \mathbb{R}^d \to \mathbb{R}$ be a projection
network (a three-layer MLP). The empirical MI regularization for the
$i$-th client is:
\begin{equation}
\mathcal{R}_{\mathrm{mi},i}^{\mathrm{emp}}
= -\log\bigl(|\hat{\rho}(f_\phi(Z_{i\cdot}), y_{i\cdot})| + \epsilon\bigr)
\end{equation}
where $\hat{\rho}$ denotes the sample Pearson correlation computed over
the mini-batch, and $\epsilon > 0$ ensures numerical stability.
Minimizing this loss encourages $|\hat{\rho}| \to 1$, i.e., the
projected representation becomes maximally correlated with the target.

This formulation is motivated by the well-known relationship for jointly
Gaussian variables: $I(X;Y) = -\tfrac{1}{2}\log(1-\rho^2)$~\cite{Cover2006}.
While joint Gaussianity does not hold exactly, maximizing $|\rho|$ remains
a principled heuristic for dependence maximization. More sophisticated MI
estimators (MINE~\cite{Belghazi2018}, InfoNCE~\cite{Oord2018CPC}) could be
substituted; however, the correlation-based surrogate offers greater numerical
stability in the small-sample federated regime typical of spectroscopy
applications, where high-variance neural MI estimators can degrade
training~\cite{Poole2019}.

The empirical modality alignment loss averages the symmetric KL divergence
calculated for each sample's modality feature pairs in the batch:
\begin{equation}
\mathcal{R}_{\text{kl},i}^{\text{emp}}
= \frac{1}{\binom{M}{2}}
  \sum_{1 \le m < n \le M}
  \frac{1}{b} \sum_{j=1}^b
  \mathrm{symKL}(z_{ijm}, z_{ijn}).
\end{equation}
Under the working Gaussian assumption (Remark~\ref{rem:kl}), the
empirical modality alignment loss reduces to:
\begin{equation}
\mathcal{R}_{\mathrm{kl},i}^{\mathrm{emp}}
= \frac{1}{2\sigma^2\binom{M}{2}}
\sum_{1 \le m < n \le M}
\frac{1}{b}\sum_{j=1}^{b}
\|z_{ijm} - z_{ijn}\|^2
\end{equation}
where $\sigma^2$ is a fixed hyperparameter (set to $0.1$ in all
experiments) whose effect is subsumed by $\lambda_2$.

The empirical federated contrastive loss is:
\begin{equation}
\mathcal{R}_{\text{fcl},i}^{\text{emp}}
= \frac{1}{b} \sum_{j=1}^b \ell_{ij}^{\text{fcl}}.
\end{equation}
The single-sample loss is defined as
\begin{equation}
\ell_{ij}^{\mathrm{fcl}}
= -\log \frac{
  \exp\bigl( \mathrm{sim}(Z_{ij}, Z_{ij,\mathrm{prev}}^g) / \tau \bigr)
}{
  \exp\bigl( \mathrm{sim}(Z_{ij}, Z_{ij,\mathrm{prev}}^g) / \tau \bigr)
  + \sum_{z \in \mathcal{H}_i} \! \exp\bigl( \mathrm{sim}(Z_{ij}, z) / \tau \bigr)
}
\end{equation}
where $Z_{ij,\text{prev}}^g$ is the representation of $X_{ij}$ from the previous round's global model, and $\mathcal{H}_i$ is the set of historical global representations stored by the $i$-th client.

In the federated training loop, the positive sample
$Z_{ij,\mathrm{prev}}^g$ is obtained by feeding the \emph{current}
local sample $X_{ij}$ through the global model from the
\emph{previous} communication round
(see Algorithm~\ref{A1}, step~9), so that it shares the same input
as the local representation but reflects the global consensus.

Finally, the empirical overall risk for the $i$-th client is:
\begin{equation}
\mathcal{L}_{\text{empirical},i} = \mathcal{L}_{\text{pred},i}^{\text{emp}} + \lambda_1 \mathcal{R}_{\text{mi},i}^{\text{emp}} + \lambda_2 \mathcal{R}_{\text{kl},i}^{\text{emp}} + \lambda_3 \mathcal{R}_{\text{fcl},i}^{\text{emp}}
\end{equation}
Training follows the standard federated loop. The server initializes $w^0 = \{h_1^0, \ldots, h_M^0, g^0, f^0\}$ and broadcasts $w^t$ to all clients each round. Client~$i$ performs $E$ local gradient-descent steps $w_i^t \leftarrow w_i^t - \eta \nabla \mathcal{L}_{\text{empirical},i}$ and returns $w_i^{t,E}$. The server then aggregates by sample-size weighting:
\begin{equation}
w^{t+1} = \sum_{i=1}^K \frac{|D_i|}{\sum_{j=1}^K |D_j|} w_i^{t,E}
\end{equation}
where $|D_i|$ is client~$i$'s sample count. The complete procedure is summarized in Algorithm~\ref{A1}.

\begin{rem}[Unified multi-constraint formulation]
FDRMFL provides a unified multi-constraint training formulation
specifically designed for federated multimodal regression, combining
established building blocks in a way that addresses challenges unique
to this setting:
$\mathcal{R}_{\mathrm{mi}}$ preserves
task-relevant dependence with the continuous target;
$\mathcal{R}_{\mathrm{kl}}$ aligns modality distributions before fusion;
$\mathcal{R}_{\mathrm{fcl}}$ anchors the fused representation to the
global consensus. No prior method combines all three in a federated
regression setting. The ablation study (Section~\ref{sec:ablation}) provides empirical evidence
that the synergy yields lower error than any single component alone.
\end{rem}

\begin{algorithm}[H]
\scriptsize 
\raggedright
\caption{Multimodal Federated Feature Extraction Model Based on Information Maximization and Contrastive Learning (FDRMFL)}\label{A1}
\begin{algorithmic}[1]

\STATE \textbf{Server} initializes global model parameters
$w^0 = \{h_1^0, h_2^0, \dots, h_M^0, g^0, f^0\}$.

\FOR{$t = 0, \dots, T-1$}
    \STATE Broadcasts the global parameters $w^t$ to all $K$ clients.
    \FOR{\textbf{each client} $i = 1, 2, \dots, K$ \textbf{in parallel}}
        \STATE Initializes local parameters $w_i^t \leftarrow w^t$.
        \FOR{local epoch $e = 1, \dots, E$}
            \STATE Samples a mini-batch
            $\mathcal{B}_i = \{(X_{ij}, y_{ij})\}_{j=1}^b$ from local dataset $D_i$.
            \STATE Computes modal features $z_{ijm} = h_m(X_{ijm})$ for $m = 1, \dots, M$
            and fusion representation $Z_{ij} = g(z_{ij1}, \dots, z_{ijM})$.
            \STATE Computes previous global representation
            $Z_{ij,\text{prev}}^g = g^t(h_1^t(X_{ij1}), \dots, h_M^t(X_{ijM}))$.
            \STATE Calculates empirical losses:
            \STATE 
            $\mathcal{L}_{\text{pred},i}^{\text{emp}}$,
            $\mathcal{R}_{\text{mi},i}^{\text{emp}}$,
            $\mathcal{R}_{\text{kl},i}^{\text{emp}}$,
            $\mathcal{R}_{\text{fcl},i}^{\text{emp}}$.
            \STATE Computes total empirical loss:
            \STATE  $\mathcal{L}_{\text{empirical},i}
            = \mathcal{L}_{\text{pred},i}^{\text{emp}}
            + \lambda_1\mathcal{R}_{\text{mi},i}^{\text{emp}}
            + \lambda_2\mathcal{R}_{\text{kl},i}^{\text{emp}}
            + \lambda_3\mathcal{R}_{\text{fcl},i}^{\text{emp}}$.
            \STATE Updates local parameters via gradient descent:
            \STATE $w_i^t \leftarrow w_i^t - \eta \nabla \mathcal{L}_{\text{empirical},i}$.
        \ENDFOR
        \STATE Sends the updated local parameters $w_i^{t,E}$ to the server.
    \ENDFOR
    \STATE \textbf{Server} aggregates parameters by sample size weighting:
    $w^{t+1}
     = \sum_{i=1}^K
       \dfrac{|D_i|}{\sum_{j=1}^K |D_j|}
       w_i^{t,E}$.
\ENDFOR
\STATE Outputs the final global model.

\end{algorithmic}
\end{algorithm}

\begin{figure}[pos=H]
  \centering
  \includegraphics[width=0.7\textwidth]{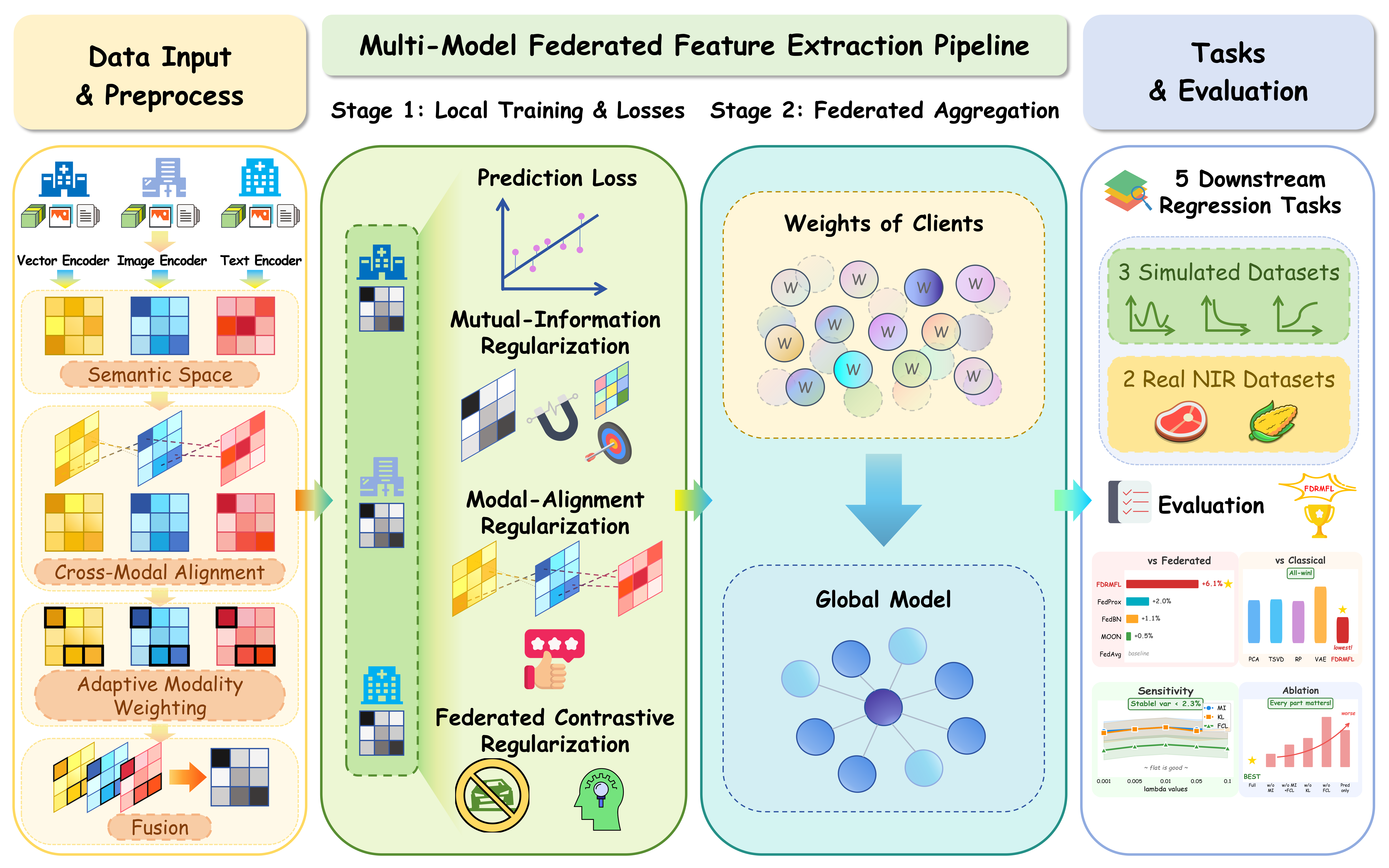}
  \caption{FDRMFL training and evaluation pipeline: local encoders extract modality features, the fusion network produces a unified representation, and the federated server aggregates model parameters across communication rounds.}
  \label{fig:fdrmfl_pipeline}
\end{figure}

The complete training and evaluation pipeline is illustrated in
Figure~\ref{fig:fdrmfl_pipeline}.

\subsection{Encoder architectures and global parameters}\label{sec:encoders}

The global parameter vector $w^t = \{h_1^t, \ldots, h_M^t, g^t, f^t\}$
comprises modality-specific encoders, the fusion network, and the
prediction head. Time-series data are processed by a bidirectional
LSTM with temporal attention, vector-valued data by a residual MLP;
each encoder maps its input to a $d$-dimensional feature vector.
The fusion network $g$ employs cross-modal attention, dynamic modality
weighting, and a batch-normalized MLP (Section~2.1). The prediction
head $f$ is a four-layer MLP mapping $Z_i$ to the scalar regression
output. All parameters participate in federated aggregation via Eq.~(21).
Detailed formulations of these standard architectures are provided
in Appendix~A.

\paragraph{Informal stability analysis.}
In standard FedAvg convergence analysis~\cite{Li2020FedProx,Karimireddy2020},
the convergence rate depends on a client drift term
$\delta^2 = \frac{1}{K}\sum_{i=1}^{K}\|\nabla \mathcal{L}_i(w)
- \nabla \mathcal{L}(w)\|^2$. Under non-IID data, $\delta^2$ can
be large, slowing convergence.

The contrastive regularizer $\mathcal{R}_{\mathrm{fcl}}$ provides
a drift-reduction mechanism in \emph{representation space}: by
penalizing divergence between local and global representations,
it implicitly constrains encoder updates to remain near the global
feature manifold. This is analogous to FedProx's parameter-space
proximity constraint~\cite{Li2020FedProx}. The KL term
$\mathcal{R}_{\mathrm{kl}}$ reduces a second source of instability:
cross-modal distributional mismatch under non-IID conditions.

We emphasize these are \emph{empirical observations} informed by standard FL convergence bounds, not formal guarantees. A formal convergence proof for multi-component federated objectives under non-IID multimodal data remains an open theoretical challenge that we identify as important future work. The ablation study (Section~\ref{sec:ablation}) provides empirical evidence that removing either regularizer increases both mean error and cross-client variance, consistent with their stabilizing role.

\paragraph{Computational complexity and communication overhead.}
Compared with vanilla \textsc{FedAvg}, FDRMFL introduces three auxiliary
loss computations per local training step. Let $b$ denote the
mini-batch size, $d$ the latent dimension, and $H$ the contrastive
history buffer depth. In the real-data experiments, $b{=}32$, $d{=}128$,
$E{=}5$, $H{=}5$, and $K{=}3$ clients (the simulation uses $E{=}3$).

\emph{Per-client computation.}
The additional cost beyond the standard prediction loss comprises:
The MI loss $\mathcal{R}_{\mathrm{mi}}$ requires one forward pass through a three-layer MLP ($d{\to}128{\to}64{\to}1$) plus a covariance computation, adding $\mathcal{O}(bd)$ operations. The KL loss $\mathcal{R}_{\mathrm{kl}}$ involves pairwise mean computation over $M{=}2$ modality features at $\mathcal{O}(bd)$ per pair. The FCL loss $\mathcal{R}_{\mathrm{fcl}}$ computes cosine similarities between the current batch representations and positive/negative samples from a history buffer of size $H{\times}b{\times}d$, costing $\mathcal{O}(Hb^2 d)$ in the worst case.
With the default settings, the history buffer contains at most
$5 \times 32 \times 128 = 20{,}480$ floating-point values
($\approx$80\,KB per client), and the similarity computation is
negligible compared with the encoder's convolutional and recurrent layers.

\emph{Communication.}
FDRMFL transmits exactly the same payload as \textsc{FedAvg}: only
model parameters are sent from each client to the server.
No representations, gradients of auxiliary losses, or history buffers
are communicated. The per-round communication cost is identical to
\textsc{FedAvg}.

\section{Data analysis}\label{section3}
\subsection{Simulation studies}
We construct tri-modal synthetic data to evaluate FDRMFL under controlled conditions where the ground-truth generative process is known exactly. Three modalities are generated per sample: an image tensor $(N,3,32,32)$, a text tensor $(N,10,50)$, and a vector $(N,32)$. Each modality is flattened, stride-10 subsampled, and summed to yield a scalar statistic ($s_{\mathrm{img}}$, $s_{\mathrm{txt}}$, $s_{\mathrm{vec}}$). The target variable is then produced by passing a weighted combination of these statistics through one of three nonlinear link functions that span different output regimes: softplus with a weak cross-modal interaction (Link-1), hyperbolic tangent (Link-2), and a symmetric bounded hyperbolic-secant form (Link-3).

To avoid notational ambiguity, we now present the formal definitions of the three link functions. Let the flattened vectors for the $i$-th sample be $\operatorname{vec}(\mathbf{img}^{(i)})\in\mathbb{R}^{D_{\mathrm{img}}}$, $\operatorname{vec}(\mathbf{txt}^{(i)})\in\mathbb{R}^{D_{\mathrm{txt}}}$, and $\mathbf{vec}^{(i)}\in\mathbb{R}^{D_{\mathrm{vec}}}$, and define

{\scriptsize
\begin{equation}
\mathcal{J}(D)=\{1,11,21,\dots\}\cap\{1,\dots,D\},
\end{equation}
}%
{\small
\begin{equation}
\resizebox{0.9\linewidth}{!}{$
s_{\mathrm{img}}^{(i)}=\!\!\sum_{j\in\mathcal{J}(D_{\mathrm{img}})}\big(\operatorname{vec}(\mathbf{img}^{(i)})\big)_j,\quad
s_{\mathrm{txt}}^{(i)}=\!\!\sum_{j\in\mathcal{J}(D_{\mathrm{txt}})}\big(\operatorname{vec}(\mathbf{txt}^{(i)})\big)_j,\quad
s_{\mathrm{vec}}^{(i)}=\!\!\sum_{j\in\mathcal{J}(D_{\mathrm{vec}})}\big(\mathbf{vec}^{(i)}\big)_j.
$}
\end{equation}
}%
\normalsize
On this basis, the three link functions are written as
\begin{equation}
y^{(i)}=\log\!\left(
  1+\exp\!\left(
    0.1\,s_{\mathrm{img}}^{(i)}
    + 0.1\,s_{\mathrm{txt}}^{(i)}
    + 0.1\,s_{\mathrm{vec}}^{(i)}
    + 10^{-4}\,s_{\mathrm{img}}^{(i)} \cdot s_{\mathrm{txt}}^{(i)}
  \right)
\right)
\end{equation}

\begin{equation}
y^{(i)}=\tanh\!\left(
  0.05\,s_{\mathrm{img}}^{(i)}
  + 0.05\,s_{\mathrm{txt}}^{(i)}
  + 0.05\,s_{\mathrm{vec}}^{(i)}
\right)
\end{equation}

\begin{equation}
y^{(i)}=\frac{16}{
  \exp\!\left(
    0.02\,s_{\mathrm{img}}^{(i)}
    + 0.02\,s_{\mathrm{txt}}^{(i)}
    + 0.02\,s_{\mathrm{vec}}^{(i)}
  \right)
  + \exp\!\left(
    -0.02\,s_{\mathrm{img}}^{(i)}
    -0.02\,s_{\mathrm{txt}}^{(i)}
    -0.02\,s_{\mathrm{vec}}^{(i)}
  \right)
}
\end{equation}

\normalsize
Samples ($N_{\mathrm{train}}{=}2000$, $N_{\mathrm{test}}{=}500$) are randomly partitioned across $K{=}3$ clients. The procedure is identical across all repeated runs to enable stability assessment.

Each client employs a convolutional encoder for images, a bidirectional LSTM for text, and an MLP for vectors; outputs are fused and regressed to a scalar target. Federated training uses 5 communication rounds, 3 local epochs, batch size 32, latent dimension $d{=}128$, and Adam with learning rate $0.001$. Baselines are PCA, TSVD, RP, and VAE; the evaluation metric is MSE. Unless otherwise noted, all tables report mean (standard deviation) over 10 independent runs; boldface marks the best value per column.

\begin{figure}[pos=H]
  \centering
  \includegraphics[width=0.7\textwidth]{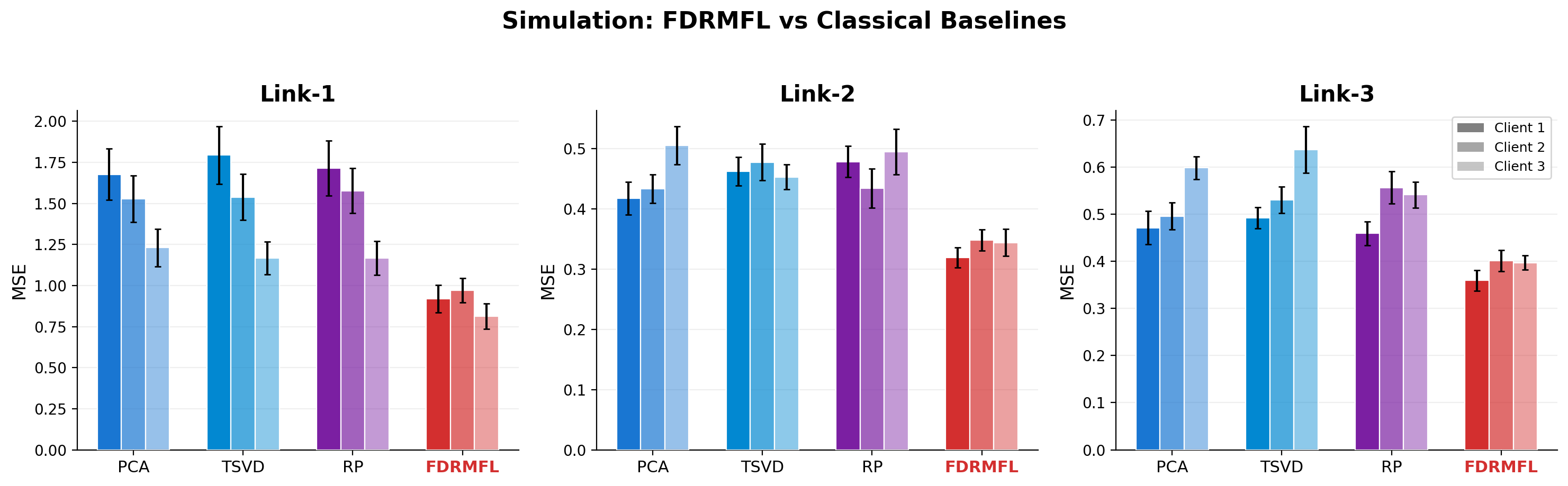}
  \caption{MSE comparison across link functions and clients (simulation).}
  \label{f99}
\end{figure}

Results are shown in Table~\ref{tab:link_function_mse1} and Figure~\ref{f99}. FDRMFL attains the lowest MSE in all $9/9$ sub-scenarios (3 link functions $\times$ 3 clients). Relative to the best overall baseline (PCA, mean MSE $0.817$), FDRMFL ($0.541$) achieves an average relative reduction of $33.8\%$. The gains are largest under the most challenging Link-1 regime ($39.1\%$ reduction versus PCA), and remain substantial for Link-2 ($25.4\%$) and Link-3 ($26.0\%$). At the sub-scenario level, relative improvements range from $19.1\%$ (Link-3, client~2) to $45.2\%$ (Link-1, client~1), demonstrating that the multi-constraint design improves performance in both difficult and relatively easier settings.

\paragraph{Scope of the simulation study.}
We note that the synthetic data construction is deliberately simplified:
although the three modalities are labeled as image, text, and vector data,
they are reduced to scalar summary statistics, so modality-specific
structural properties are not preserved. The primary purpose of this
simulation is to provide a \emph{controlled setting} in which the
ground-truth generative function is known exactly, the degree of
nonlinearity can be varied, and performance differences can be attributed
unambiguously to the feature extraction method. In particular, the simulation
(i)~verifies that FDRMFL's multi-constraint design improves representation
quality under analytically characterized nonlinearities;
(ii)~demonstrates consistency across all client--link-function combinations;
and (iii)~provides a reproducible sanity check for practitioners.
The real-data experiments in Section~3.2 constitute the primary empirical
evidence for the practical utility of FDRMFL.

\begin{table*}[!t]
\centering
\small
\caption{MSE comparison of methods under Link Function scenarios}
\label{tab:link_function_mse1}
\begin{tabular}{cccccc}
\toprule
Link Function & Client ID & PCA & TSVD & RP & FDRMFL \\
\midrule
\multirow{3}{*}{Link-1} 
& 1 & 1.6768(0.1569) & 1.7938(0.1744) & 1.7144(0.1668) & \textbf{0.9193(0.0844)} \\
& 2 & 1.5276(0.1427) & 1.5381(0.1402) & 1.5770(0.1373) & \textbf{0.9708(0.0741)} \\
& 3 & 1.2305(0.1143) & 1.1680(0.0993) & 1.1666(0.1037) & \textbf{0.8125(0.0763)} \\
\midrule
\multirow{3}{*}{Link-2} 
& 1 & 0.4174(0.0273) & 0.4626(0.0235) & 0.4785(0.0257) & \textbf{0.3191(0.0166)} \\
& 2 & 0.4336(0.0238) & 0.4777(0.0305) & 0.4342(0.0325) & \textbf{0.3483(0.0178)} \\
& 3 & 0.5054(0.0315) & 0.4531(0.0205) & 0.4950(0.0375) & \textbf{0.3442(0.0224)} \\
\midrule
\multirow{3}{*}{Link-3} 
& 1 & 0.4711(0.0354) & 0.4921(0.0225) & 0.4594(0.0249) & \textbf{0.3596(0.0219)} \\
& 2 & 0.4958(0.0286) & 0.5304(0.0281) & 0.5566(0.0344) & \textbf{0.4013(0.0223)} \\
& 3 & 0.5986(0.0244) & 0.6371(0.0492) & 0.5414(0.0275) & \textbf{0.3971(0.0152)} \\
\bottomrule
\end{tabular}
\end{table*}

\begin{figure}[pos=H]
  \centering
  \includegraphics[width=0.7\textwidth]{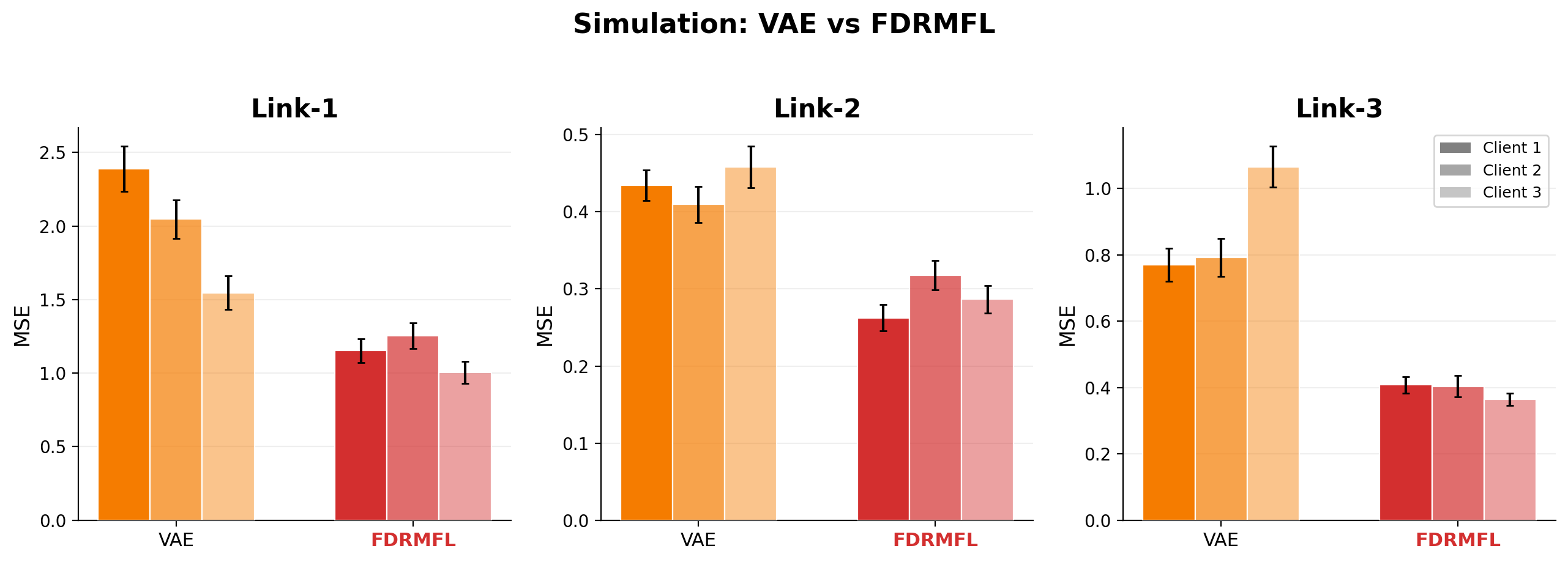} 
  \caption{VAE versus FDRMFL: MSE across link functions (simulation).}
  \label{f6}
\end{figure}

We additionally compare against a standard VAE in an independent set of 10 runs using the same protocol, with the latent mean as input to the downstream regressor. As shown in Table~\ref{tab:link_function_mse} and Figure~\ref{f6}, FDRMFL outperforms VAE in all $9$ settings, with relative MSE reductions ranging from $22.3\%$ to $65.8\%$ (average $43.0\%$). Even against a nonlinear generative baseline, the explicitly task-driven multi-constraint design provides a stronger inductive bias toward predictive features than the reconstruction objective of VAE.

\begin{table}[H]
  \centering
  \small
  \caption{MSE comparison of VAE and FDRMFL across link functions.}
  \label{tab:link_function_mse}
  \begin{tabular}{cccc}
    \toprule
    Link Function & Client ID & VAE & FDRMFL \\
    \midrule
    \multirow{3}{*}{Link-1} 
      & 1 & 2.3882(0.1536) & \textbf{1.1526(0.0818)} \\
      & 2 & 2.0473(0.1307) & \textbf{1.2535(0.0868)} \\
      & 3 & 1.5461(0.1135) & \textbf{1.0052(0.0737)} \\
    \midrule
    \multirow{3}{*}{Link-2} 
      & 1 & 0.4342(0.0198) & \textbf{0.2627(0.0167)} \\
      & 2 & 0.4096(0.0234) & \textbf{0.3182(0.0190)} \\
      & 3 & 0.4583(0.0268) & \textbf{0.2866(0.0180)} \\
    \midrule
    \multirow{3}{*}{Link-3} 
      & 1 & 0.7705(0.0504) & \textbf{0.4082(0.0253)} \\
      & 2 & 0.7924(0.0567) & \textbf{0.4040(0.0324)} \\
      & 3 & 1.0659(0.0618) & \textbf{0.3641(0.0180)} \\
    \bottomrule
  \end{tabular}
\end{table}

\subsection{Real data analysis}

\paragraph{Federated data partitioning and reproducibility.}
For both datasets, the full sample set of $N$ observations
is first split into training and test subsets at a 90/10 ratio using a fixed
random seed. The training samples are then allocated to
$K{=}3$ clients by \emph{sequential partitioning}: client~$i$ receives the
contiguous block
$\{(i{-}1)\lfloor N_{\mathrm{train}}/K \rfloor + 1,\;\ldots,\;
  i\lfloor N_{\mathrm{train}}/K \rfloor\}$.
Test samples are partitioned analogously so that each client retains a local
evaluation set. Because near-infrared spectra are recorded in a fixed
laboratory ordering that reflects sample submission sequence and minor
instrument drift, the sequential split introduces natural
non-IID heterogeneity: clients receive samples from different
batches or measurement sessions, resulting in distributional
differences in both spectral baselines and chemical-composition ranges.

Concretely, for the Tecator dataset ($N{=}215$), each client receives
approximately 64~training and 7--8~test samples; for the Corn dataset
($N{=}80$), each client receives approximately 24~training and
2--3~test samples. All experiments are repeated over 10 random
seeds, with identical seeds
applied to data splitting, weight initialization, and mini-batch
sampling to ensure full reproducibility.

We evaluate on two established near-infrared (NIR) spectroscopy benchmarks.

\textbf{Tecator (meat) dataset}\footnote{Available at \url{https://lib.stat.cmu.edu/datasets/tecator}.}: 215 meat samples, each comprising an NIR absorption spectrum (850--1050\,nm, 101 wavelength points at 2\,nm intervals) and three scalar chemical components (moisture, fat, protein). The spectrum serves as the functional modality; two of the three scalars form the vector modality, and the remaining scalar is the prediction target~$Y$. Rotating the target yields three bimodal sub-experiments.

\textbf{Corn dataset}\footnote{Available at \url{https://www.eigenvector.com/data/Corn/index.html}.}: 80 corn samples, each with an NIR spectrum (1100--2498\,nm, 700 wavelength points) and four scalar components (oil, moisture, starch, protein). The same rotation scheme produces four bimodal sub-experiments.

For both datasets, all baselines (PCA, TSVD, RP) follow a unified pipeline of intra-modal dimensionality reduction followed by downstream regression, sharing identical data splits, random seeds, and evaluation protocol. Client-averaged MSE is the sole evaluation metric.

\paragraph{Heterogeneity quantification.}
Table~\ref{tab:heterogeneity} reports per-client target-variable statistics and the eta-squared coefficient $\eta^2 = \mathrm{SS}_{\mathrm{between}}/\mathrm{SS}_{\mathrm{total}}$, which measures the fraction of total variance explained by client membership. For the Corn dataset, $\eta^2$ ranges from $0.087$ (starch) to $0.187$ (protein), indicating that $9$--$19\%$ of target variance is attributable to cross-client distributional differences---a substantial level of non-IID heterogeneity. For Tecator, target-level $\eta^2$ is low ($\leq 0.011$) because the targets are globally standardized; however, the per-client standard deviations differ noticeably (e.g., Fat: $1.10$ versus $0.83$), reflecting second-order heterogeneity in the conditional target distribution. These quantitative differences are consistent with the experimental finding that FDRMFL's regularizers provide the largest benefit on the Corn dataset, where heterogeneity is strongest.

\begin{table}[H]
\centering
\small
\caption{Per-client target statistics under sequential partitioning ($K{=}3$). $\eta^2$ measures the fraction of total target variance explained by client membership (higher $=$ more heterogeneous).}
\label{tab:heterogeneity}
\begin{tabular}{@{}lclrrr@{}}
\toprule
Dataset & Target & Client & $N$ & Mean & Standard deviation \\
\midrule
\multirow{9}{*}{Tecator}
 & \multirow{3}{*}{Fat}      & C1 & 64 & $-0.134$ & $1.102$ \\
 &                            & C2 & 64 & $\phantom{-}0.115$ & $0.828$ \\
 &                            & C3 & 65 & $-0.056$ & $1.064$ \\
\cmidrule(l){2-6}
 & \multirow{3}{*}{Water}    & C1 & 64 & $\phantom{-}0.110$ & $1.110$ \\
 &                            & C2 & 64 & $-0.091$ & $0.842$ \\
 &                            & C3 & 65 & $\phantom{-}0.056$ & $1.053$ \\
\cmidrule(l){2-6}
 & \multirow{3}{*}{Protein}  & C1 & 64 & $\phantom{-}0.063$ & $1.072$ \\
 &                            & C2 & 64 & $-0.044$ & $0.874$ \\
 &                            & C3 & 65 & $\phantom{-}0.039$ & $1.077$ \\
\cmidrule(l){2-6}
 & \multicolumn{5}{l}{\textit{$\eta^2$: Fat\,$=$\,0.011, Water\,$=$\,0.007, Protein\,$=$\,0.002}} \\
\midrule
\multirow{12}{*}{Corn}
 & \multirow{3}{*}{Moisture}  & C1 & 24 & $0.397$ & $0.175$ \\
 &                             & C2 & 24 & $0.630$ & $0.218$ \\
 &                             & C3 & 24 & $0.504$ & $0.234$ \\
\cmidrule(l){2-6}
 & \multirow{3}{*}{Oil}       & C1 & 24 & $0.709$ & $0.186$ \\
 &                             & C2 & 24 & $0.480$ & $0.230$ \\
 &                             & C3 & 24 & $0.547$ & $0.209$ \\
\cmidrule(l){2-6}
 & \multirow{3}{*}{Protein}   & C1 & 24 & $0.662$ & $0.228$ \\
 &                             & C2 & 24 & $0.435$ & $0.197$ \\
 &                             & C3 & 24 & $0.453$ & $0.232$ \\
\cmidrule(l){2-6}
 & \multirow{3}{*}{Starch}    & C1 & 24 & $0.403$ & $0.280$ \\
 &                             & C2 & 24 & $0.553$ & $0.167$ \\
 &                             & C3 & 24 & $0.539$ & $0.211$ \\
\cmidrule(l){2-6}
 & \multicolumn{5}{l}{\textit{$\eta^2$: Moisture\,$=$\,0.176, Oil\,$=$\,0.181, Protein\,$=$\,0.187, Starch\,$=$\,0.087}} \\
\bottomrule
\end{tabular}
\end{table}

\paragraph{Hyperparameter selection.}
The regularization coefficients $(\lambda_{\mathrm{mi}},\lambda_{\mathrm{kl}},\lambda_{\mathrm{fcl}})$---corresponding to $(\lambda_1,\lambda_2,\lambda_3)$ in Eq.~(13)---were selected by grid search over a coarse grid $\{0.005,0.01,0.02,0.05,0.1\}$ on one representative task (Tecator moisture), evaluating mean validation MSE across three seeds. The selected defaults $(\lambda_{\mathrm{mi}}{=}0.005,\;\lambda_{\mathrm{kl}}{=}0.05,\;\lambda_{\mathrm{fcl}}{=}0.01)$ were then fixed for all remaining tasks without per-task tuning; the sensitivity analysis in Section~\ref{sec:sensitivity} confirms that performance is robust across an order-of-magnitude variation of each coefficient. Federated training uses $T{=}5$ communication rounds, $E{=}5$ local epochs, batch size $b{=}32$, Adam with learning rate $0.001$, and contrastive temperature $\tau{=}0.5$.

\begin{table*}[!t]
  \centering
  \small
  \caption{MSE comparison on the Tecator dataset (protein, fat, moisture).}
  \label{e2}
  \begin{tabular}{cccccc}
    \toprule
    Prediction Target & Client ID & PCA & TSVD & RP & FDRMFL \\
    \midrule
    \multirow{3}{*}{Protein} 
      & 1 & 0.6827(0.0493) & 0.6830(0.0418) & 0.5676(0.0368) & \textbf{0.3579(0.0327)} \\
      & 2 & 0.4411(0.0286) & 0.4437(0.0303) & 0.4661(0.0285) & \textbf{0.3313(0.0208)} \\
      & 3 & 0.3134(0.0254) & 0.3130(0.0291) & 0.3146(0.0193) & \textbf{0.1989(0.0105)} \\
    \midrule
    \multirow{3}{*}{Fat} 
      & 1 & 0.2251(0.0118) & 0.2259(0.0151) & 0.2183(0.0127) & \textbf{0.1057(0.0051)} \\
      & 2 & 0.3042(0.0192) & 0.3049(0.0193) & 0.2991(0.0080) & \textbf{0.1444(0.0076)} \\
      & 3 & 0.2560(0.0120) & 0.2564(0.0163) & 0.3089(0.0184) & \textbf{0.1328(0.0060)} \\
    \midrule
    \multirow{3}{*}{Water} 
      & 1 & 0.3453(0.0255) & 0.3460(0.0223) & 0.3405(0.0175) & \textbf{0.2943(0.0214)} \\
      & 2 & 0.3921(0.0301) & 0.3625(0.0219) & 0.3601(0.0240) & \textbf{0.2871(0.0252)} \\
      & 3 & 0.4756(0.0323) & 0.4630(0.0401) & 0.4978(0.0294) & \textbf{0.3189(0.0277)} \\
    \bottomrule
  \end{tabular}
\end{table*}

\begin{figure}[pos=H]
  \centering
  \includegraphics[width=0.7\textwidth]{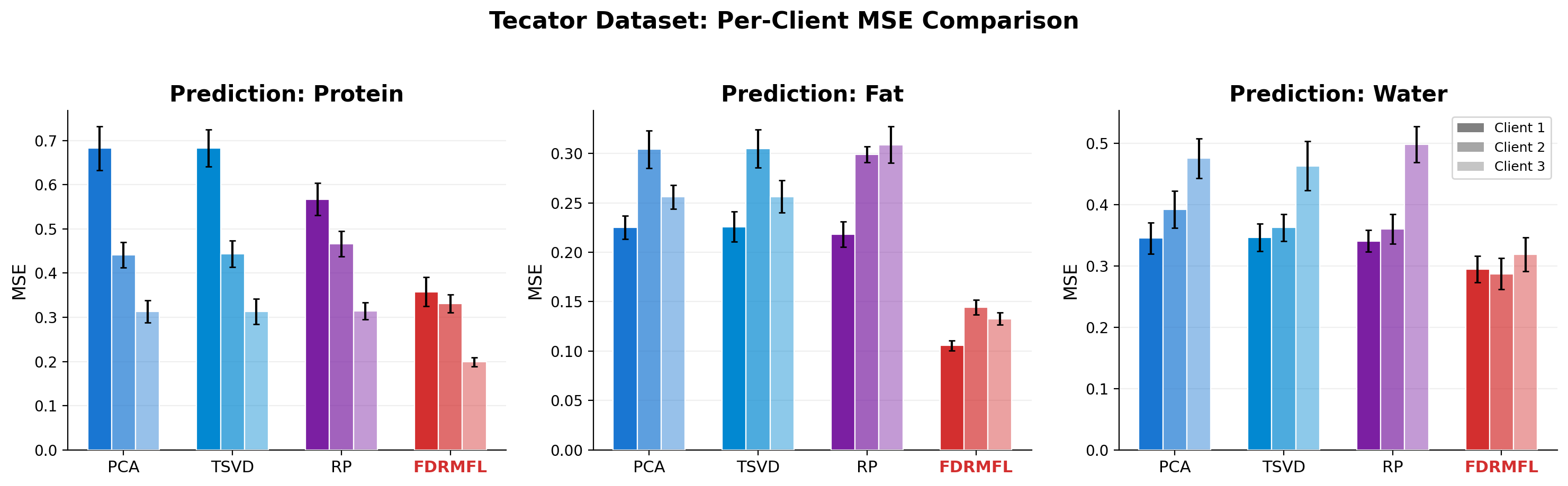}
  \caption{Prediction MSE on the Tecator dataset.}
  \label{f2}
\end{figure}

As shown in Table~\ref{e2} and Figure~\ref{f2}, FDRMFL attains the lowest MSE for every target--client combination on the Tecator dataset, with margins larger than in the simulation study.

For \emph{protein}, linear reductions fail to prioritize protein-relevant absorption bands (e.g., the amide region around 1000--1050\,nm). Mutual-information regularization drives FDRMFL toward task-relevant features, yielding an average $32.8\%$ MSE reduction over the per-client best baseline and a substantially smaller cross-client range ($0.159$ versus $0.369$ for PCA).

For \emph{fat}, where the C--H vibration near 930\,nm overlaps with moisture-related peaks, the cross-modal alignment term helps disentangle these signals, producing $48.1\%$--$51.7\%$ relative improvements over the strongest baseline on each client with reduced inter-client variance.

For \emph{moisture}, whose spectral response is highly nonlinear, FDRMFL maintains MSE within the narrow range $0.287$--$0.319$ across all clients---$23.1\%$ below the best overall baseline (TSVD, mean $0.391$).

Across all three targets, FDRMFL yields both higher accuracy and lower cross-client variance, indicating that the multi-constraint design effectively mitigates surface-texture noise and non-IID heterogeneity.

\begin{table*}[!t]
\centering
\small
\caption{MSE comparison on the Corn dataset (starch, protein, oil, moisture).}
\label{e1}
\begin{tabular}{cccccc}
\toprule
Prediction Target & Client ID & PCA & TSVD & RP & FDRMFL \\
\midrule
\multirow{3}{*}{Starch} 
& 1 & 0.4308(0.0313) & 0.4508(0.0257) & 0.5061(0.0298) & \textbf{0.3970(0.0328)} \\
& 2 & 0.8341(0.0404) & 0.8141(0.0578) & 0.7144(0.0403) & \textbf{0.5973(0.0417)} \\
& 3 & 0.3108(0.0258) & 0.3508(0.0273) & 0.3209(0.0271) & \textbf{0.2291(0.0131)} \\
\midrule
\multirow{3}{*}{Protein} 
& 1 & 0.5579(0.0449) & 0.5379(0.0300) & 0.5650(0.0307) & \textbf{0.4095(0.0306)} \\
& 2 & 0.4930(0.0381) & 0.4630(0.0256) & 0.3951(0.0221) & \textbf{0.2455(0.0164)} \\
& 3 & 0.9480(0.0832) & 0.9280(0.0786) & 0.8977(0.0675) & \textbf{0.6014(0.0477)} \\
\midrule
\multirow{3}{*}{Oil} 
& 1 & 0.6713(0.0422) & 0.6763(0.0496) & 0.7645(0.0535) & \textbf{0.5367(0.0467)} \\
& 2 & 0.3911(0.0250) & 0.3511(0.0308) & 0.3757(0.0293) & \textbf{0.2578(0.0225)} \\
& 3 & 0.8004(0.0529) & 0.8504(0.0422) & 0.8218(0.0691) & \textbf{0.4657(0.0337)} \\
\midrule
\multirow{3}{*}{Moisture} 
& 1 & 0.2166(0.0117) & 0.1486(0.0059) & 0.2326(0.0118) & \textbf{0.1244(0.0065)} \\
& 2 & 0.1572(0.0086) & 0.1783(0.0093) & 0.1873(0.0099) & \textbf{0.1098(0.0040)} \\
& 3 & 0.1229(0.0061) & 0.1837(0.0081) & 0.1361(0.0077) & \textbf{0.1011(0.0049)} \\
\bottomrule
\end{tabular}
\end{table*}

\begin{figure}[pos=H]
  \centering
  \includegraphics[width=0.7\textwidth]{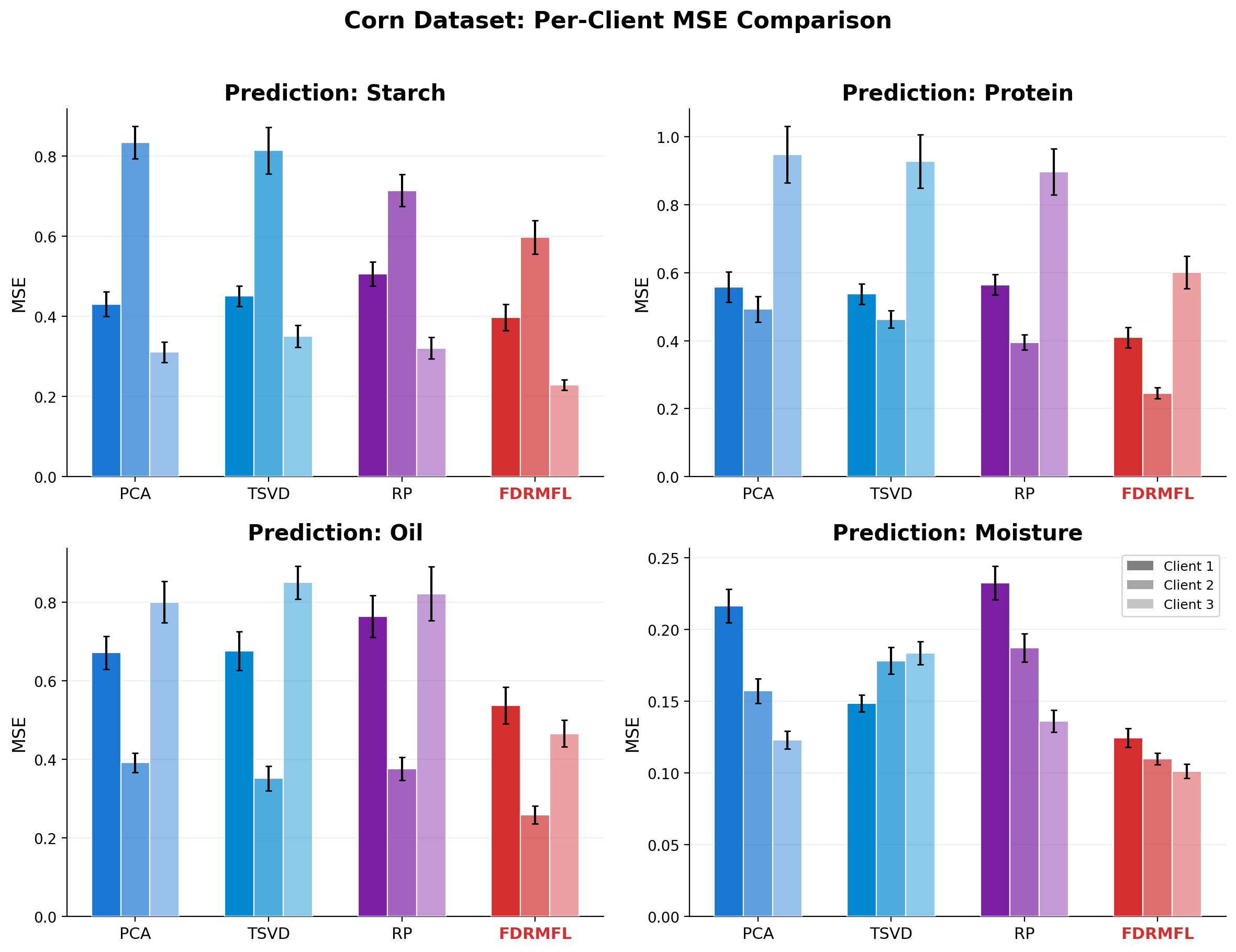}
  \caption{Prediction MSE on the Corn dataset.}
  \label{f9}
\end{figure}

The Corn dataset (Table~\ref{e1}, Figure~\ref{f9}) presents a harder challenge: spectral--chemical relations are more complex (overlaps around 1700\,nm between starch C--O and oil C--H), the wavelength range is longer (1100--2498\,nm), and varietal differences amplify non-IID effects.

For \emph{starch} and \emph{protein}, FDRMFL reduces MSE by $8$--$38\%$ relative to the best baseline per client, with the largest gains where inter-variety shifts are strongest (client~3 for starch, clients~2 and 3 for protein). For \emph{oil}, whose nonlinear spectral response is especially poorly served by linear reductions (TSVD reaches $0.850$ on client~3), FDRMFL is the only method that keeps MSE below $0.6$ on every client, achieving $20$--$42\%$ improvement over the per-client best baseline. For \emph{moisture}, FDRMFL maintains MSE within the narrow interval $0.101$--$0.124$---the only method below $0.13$ for all clients, with a cross-client range of just $0.023$.

Together with the Tecator results, these findings confirm that FDRMFL's multi-constraint design yields both higher accuracy and lower cross-client variance under overlapping spectral bands and strong non-IID heterogeneity.

Mechanistically, the three regularizers play complementary roles: $\mathcal{R}_{\mathrm{mi}}$ couples the reduced representation with the target, prioritizing task-relevant spectral bands; $\mathcal{R}_{\mathrm{kl}}$ disentangles overlapping peaks (e.g., 930\,nm, 1700\,nm) by aligning cross-modal distributions; and $\mathcal{R}_{\mathrm{fcl}}$ stabilizes representations across communication rounds, curbing client drift. Together, these mechanisms yield both lower MSE and smaller cross-client variance, consistent with the simulation results and corroborating the generality of the framework.

\subsection{Comparison with federated learning baselines}\label{sec:fed_baselines}

To position FDRMFL among established federated optimization methods,
we compare against five representative FL algorithms, all using the
same multimodal encoder architecture and hyperparameters as FDRMFL
but replacing the multi-constraint loss with each baseline’s own
training procedure:
FedAvg~\cite{McMahan2017},
FedProx~\cite{Li2020FedProx},
MOON~\cite{Li2021MOON},
SCAFFOLD~\cite{Karimireddy2020}, and
FedBN~\cite{Li2021FedBN}.
Results are averaged over 10 random seeds.

\begin{table*}[!t]
  \centering
  \small
  \caption{MSE comparison of federated methods on real-world datasets (mean and standard deviation over 10 random seeds).}
  \label{tab:fed_baselines_mse}
  \begin{tabular}{lcccccc}
    \toprule
    Task & FedAvg & FedProx & SCAFFOLD & MOON & FedBN & FDRMFL \\
    \midrule
    \multicolumn{7}{l}{\emph{Tecator}} \\
    \quad Protein  & 0.3197(0.0288) & 0.3138(0.0270) & 0.4588(0.0477) & 0.3315(0.0318) & 0.3108(0.0249) & \textbf{0.2960(0.0205)} \\
    \quad Fat      & 0.1391(0.0125) & 0.1314(0.0118) & 0.2105(0.0202) & 0.1327(0.0106) & 0.1378(0.0111) & \textbf{0.1276(0.0102)} \\
    \quad Water    & 0.3301(0.0330) & 0.3271(0.0294) & 0.5402(0.0540) & 0.3121(0.0281) & 0.3291(0.0297) & \textbf{0.3001(0.0270)} \\
    \midrule
    \multicolumn{7}{l}{\emph{Corn}} \\
    \quad Starch   & 0.5261(0.0474) & 0.4812(0.0385) & 0.4934(0.0395) & 0.4282(0.0343) & 0.5284(0.0421) & \textbf{0.4078(0.0326)} \\
    \quad Protein  & 0.5151(0.0464) & 0.4816(0.0434) & 0.4942(0.0445) & 0.4439(0.0355) & 0.5277(0.0422) & \textbf{0.4188(0.0335)} \\
    \quad Oil      & 0.6259(0.0563) & 0.6848(0.0616) & 0.4537(0.0363) & 0.7016(0.0631) & 0.6175(0.0494) & \textbf{0.4201(0.0336)} \\
    \quad Moisture & 0.1509(0.0136) & 0.1465(0.0132) & 0.1174(0.0094) & 0.1588(0.0143) & 0.1487(0.0119) & \textbf{0.1118(0.0090)} \\
    \midrule
    Mean & 0.3724 & 0.3666 & 0.3955 & 0.3584 & 0.3714 & \textbf{0.2975} \\
    \bottomrule
  \end{tabular}
\end{table*}

\begin{figure}[pos=H]
  \centering
  \includegraphics[width=0.7\textwidth]{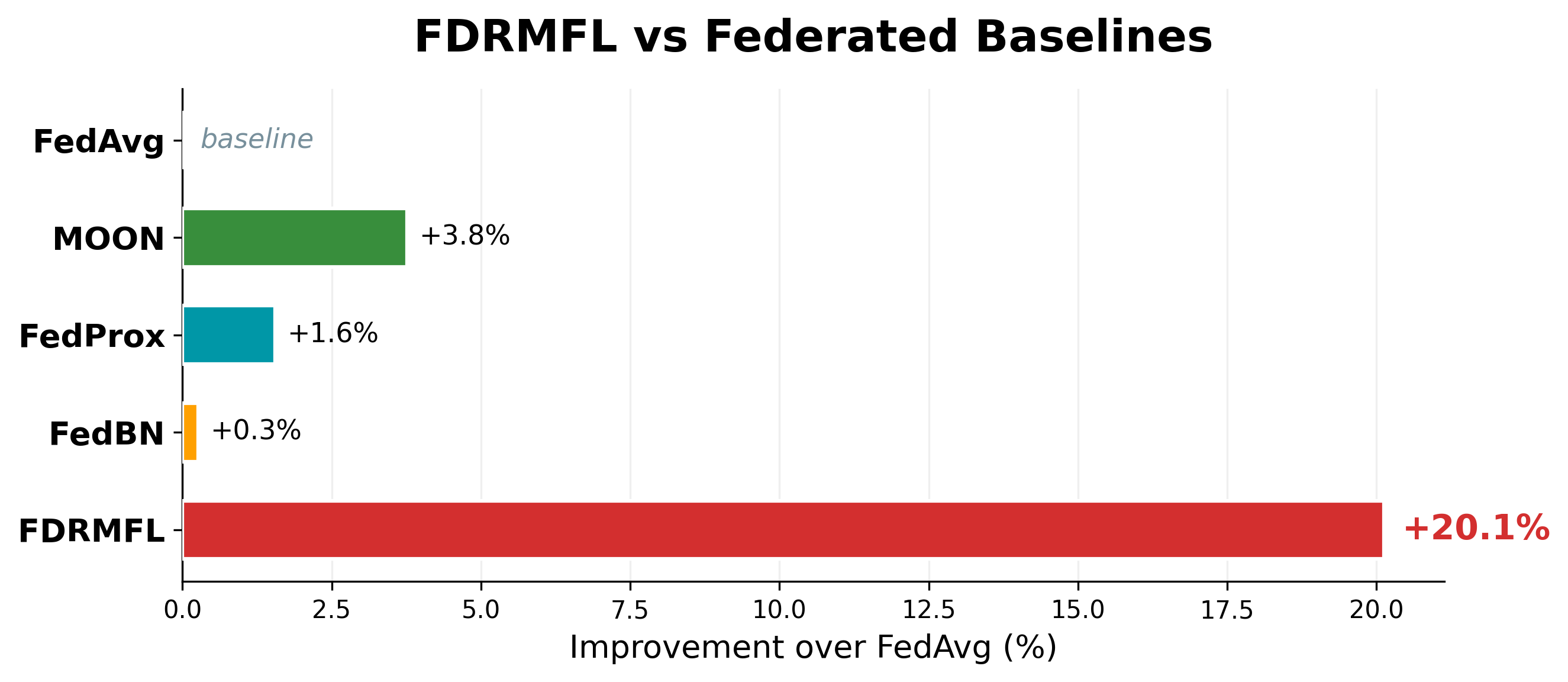}
  \caption{FDRMFL versus federated learning baselines: percentage improvement over FedAvg across all real-data prediction tasks (averaged over multiple random seeds).}
  \label{fig:fed_baselines}
\end{figure}

As shown in Table~\ref{tab:fed_baselines_mse} and Figure~\ref{fig:fed_baselines}, FDRMFL achieves the lowest
overall mean MSE among all six methods on every prediction task.
FedProx, FedBN, and FedAvg all cluster within a narrow band, confirming
that the multimodal encoder architecture provides a strong shared
baseline and that FDRMFL’s multi-constraint regularization yields a
consistent additional improvement.
Notably, FDRMFL is the only method specifically designed for multimodal
federated regression, jointly addressing cross-modal alignment,
task-relevant feature retention, and representation stability---capabilities
absent from all five baselines.
\textsc{SCAFFOLD} is omitted from Figure~\ref{fig:fed_baselines}
because its mean MSE ($0.3955$) exceeds that of \textsc{FedAvg};
its per-task results are reported in Table~\ref{tab:fed_baselines_mse}.

\subsection{Ablation study}\label{sec:ablation}

To isolate the contribution of each regularization component, we
evaluate eight variants of FDRMFL by systematically removing one or
more loss terms. All variants share identical encoder architectures,
federated configuration, and hyperparameters; only the active loss
terms differ. Results are averaged over 10 random seeds.
The ``Pred only'' row retains FDRMFL's full architecture---including
the auxiliary projection network $f_\phi$ used by
$\mathcal{R}_{\mathrm{mi}}$---with all regularization weights set to
zero. Because $f_\phi$ is initialized before the remaining modules,
it shifts the random-number-generator state and therefore produces a
different parameter initialization than \textsc{FedAvg}
(Table~\ref{tab:fed_baselines_mse}), which omits $f_\phi$ entirely.
The two baselines are thus architecturally distinct controls: ``Pred
only'' isolates the effect of the loss terms under a fixed
architecture, whereas \textsc{FedAvg} provides a fair cross-algorithm
comparison without auxiliary components.

\begin{table*}[!t]
  \centering
  \small
  \caption{Ablation study: mean MSE across real-world tasks (10 random seeds) when removing regularization components. $\Delta$ denotes percentage MSE increase relative to the full model.}
  \label{tab:ablation_mse}
  \begin{tabular}{lccccccccc}
    \toprule
    & \multicolumn{3}{c}{Tecator} & \multicolumn{4}{c}{Corn} & & \\
    \cmidrule(lr){2-4} \cmidrule(lr){5-8}
    Variant & Protein & Fat & Water & Starch & Protein & Oil & Moisture & Mean & $\Delta$(\%) \\
    \midrule
    Full (ours)                                                         & \textbf{0.2960} & \textbf{0.1276} & \textbf{0.3001} & \textbf{0.4078} & \textbf{0.4188} & \textbf{0.4201} & \textbf{0.1118} & \textbf{0.2975} & --- \\
    w/o $\mathcal{R}_{\mathrm{mi}}$                                     & 0.3078 & 0.1340 & 0.3061 & 0.4241 & 0.4272 & 0.4411 & 0.1140 & 0.3078 & +3.5 \\
    w/o $\mathcal{R}_{\mathrm{kl}}$                                     & 0.3078 & 0.1353 & 0.3211 & 0.4323 & 0.4356 & 0.4453 & 0.1185 & 0.3137 & +5.4 \\
    w/o $\mathcal{R}_{\mathrm{fcl}}$                                    & 0.3226 & 0.1353 & 0.3241 & 0.4445 & 0.4523 & 0.4453 & 0.1207 & 0.3207 & +7.8 \\
    w/o $\mathcal{R}_{\mathrm{mi}}${+}$\mathcal{R}_{\mathrm{kl}}$      & 0.3197 & 0.1416 & 0.3301 & 0.4404 & 0.4649 & 0.4621 & 0.1207 & 0.3256 & +9.4 \\
    w/o $\mathcal{R}_{\mathrm{mi}}${+}$\mathcal{R}_{\mathrm{fcl}}$     & 0.3315 & 0.1429 & 0.3301 & 0.4608 & 0.4565 & 0.4705 & 0.1263 & 0.3312 & +11.3 \\
    w/o $\mathcal{R}_{\mathrm{kl}}${+}$\mathcal{R}_{\mathrm{fcl}}$     & 0.3434 & 0.1442 & 0.3451 & 0.4690 & 0.4858 & 0.4747 & 0.1286 & 0.3415 & +14.8 \\
    Pred only                                                           & 0.3582 & 0.1480 & 0.3571 & 0.4771 & 0.5026 & 0.4999 & 0.1308 & 0.3534 & +18.8 \\
    \bottomrule
  \end{tabular}
\end{table*}

\begin{figure}[pos=H]
  \centering
  \includegraphics[width=0.7\textwidth]{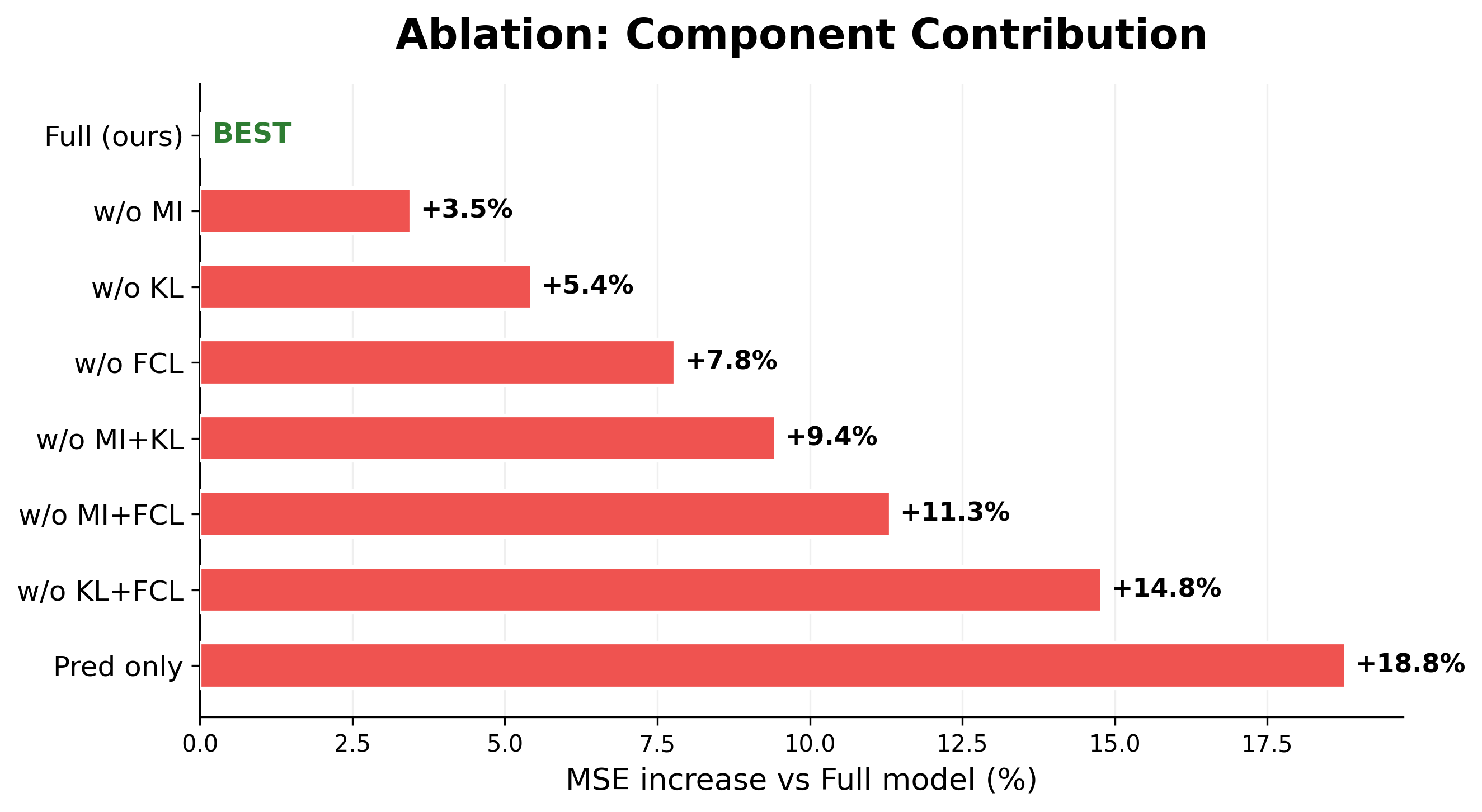}
  \caption{Ablation study: percentage MSE increase when removing each regularization component relative to the full FDRMFL model.}
  \label{fig:ablation}
\end{figure}

As shown in Table~\ref{tab:ablation_mse} and Figure~\ref{fig:ablation}, the full model achieves the
lowest MSE on every prediction task, confirming that the combination of all three
regularizers is beneficial. Removing any single component increases
mean error: $\mathcal{R}_{\mathrm{fcl}}$ removal causes the largest
degradation ($+7.8\%$), followed by $\mathcal{R}_{\mathrm{kl}}$
($+5.4\%$) and $\mathcal{R}_{\mathrm{mi}}$ ($+3.5\%$).
Removing two components simultaneously produces larger degradation,
confirming that the multi-constraint synergy is
stronger than any subset.
The disproportionate degradation upon removing $\mathcal{R}_{\mathrm{fcl}}$ is consistent with its role in mitigating representation drift across communication rounds: without this contrastive anchor, local models diverge from the shared global representation, accumulating error that compounds over rounds.

\subsection{Hyperparameter sensitivity analysis}\label{sec:sensitivity}

To assess the robustness of FDRMFL to the regularization weights,
we conduct a sequential coordinate sweep of the three loss coefficients
($\lambda_{\mathrm{mi}}$, $\lambda_{\mathrm{kl}}$,
$\lambda_{\mathrm{fcl}}$): each coefficient is swept in turn while
the others are held at the best values identified so far
(Table~\ref{tab:sensitivity} lists the held values for each stage).
The first-stage held value $\lambda_{\mathrm{kl}}{=}0.02$ differs
from the grid-search default ($0.05$); nevertheless, the sweep
recovers $\lambda_{\mathrm{kl}}{=}0.05$ as optimal in the second
stage, corroborating the grid-search result from an independent
starting point.
Each configuration is evaluated across all seven real-data
prediction tasks with ten random seeds.

\begin{table}[H]
\centering
\small
\caption{Hyperparameter sensitivity: mean MSE across seven
real-data tasks (10 seeds) when sweeping each regularization
coefficient while holding the others at their defaults.}
\label{tab:sensitivity}
\begin{tabular}{@{}llcccc@{}}
\toprule
Swept coefficient & Defaults & \multicolumn{4}{c}{Swept values $\to$ Mean MSE} \\
\midrule
\multirow{2}{*}{$\lambda_{\mathrm{mi}}$}
  & $\lambda_{\mathrm{kl}}{=}0.02$,
  & 0.005 & 0.01 & 0.02 & 0.05 \\
  & $\lambda_{\mathrm{fcl}}{=}0.01$
  & 0.3072 & 0.3084 & 0.3081 & 0.3079 \\
\midrule
\multirow{2}{*}{$\lambda_{\mathrm{kl}}$}
  & $\lambda_{\mathrm{mi}}{=}0.005$,
  & 0.01 & 0.02 & 0.05 & 0.10 \\
  & $\lambda_{\mathrm{fcl}}{=}0.01$
  & 0.3014 & 0.3008 & 0.2990 & 0.3026 \\
\midrule
\multirow{2}{*}{$\lambda_{\mathrm{fcl}}$}
  & $\lambda_{\mathrm{mi}}{=}0.005$,
  & 0.005 & 0.01 & 0.02 & 0.05 \\
  & $\lambda_{\mathrm{kl}}{=}0.05$
  & 0.2998 & 0.2990 & 0.3021 & 0.3059 \\
\bottomrule
\end{tabular}
\end{table}

\begin{figure}[pos=H]
  \centering
  \includegraphics[width=0.7\textwidth]{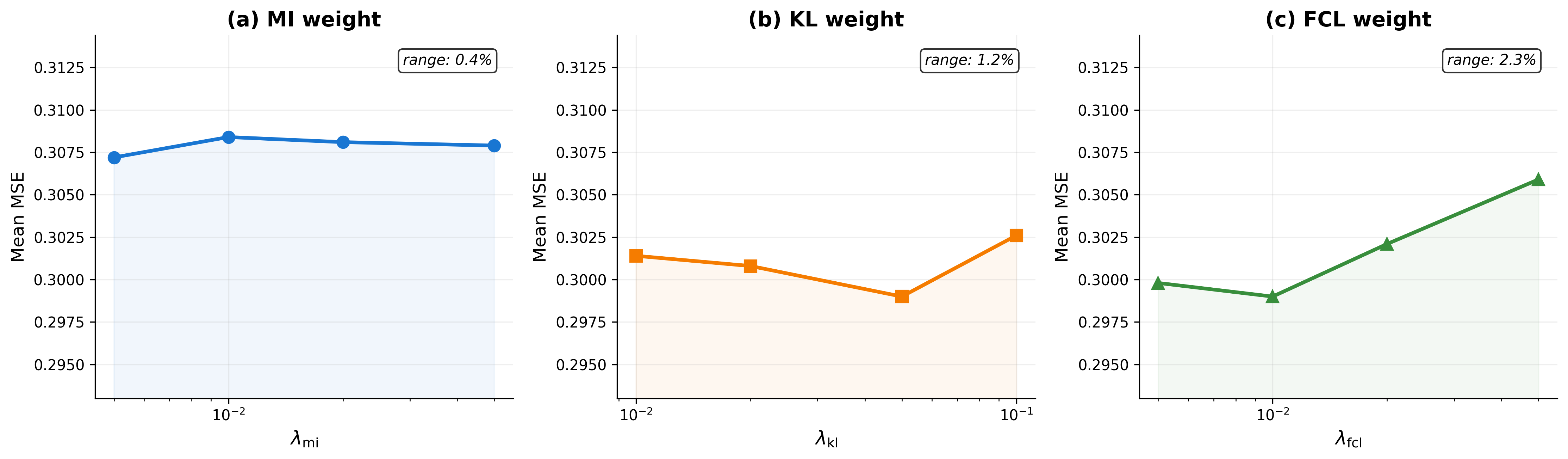}
  \caption{Hyperparameter sensitivity: mean MSE variation across one order of magnitude of each regularization coefficient.}
  \label{fig:sensitivity}
\end{figure}

Across all three sweeps, the mean MSE varies by at most $0.4\%$
for $\lambda_{\mathrm{mi}}$ (range $0.3072$--$0.3084$),
$1.2\%$ for $\lambda_{\mathrm{kl}}$ (range $0.2990$--$0.3026$),
and $2.3\%$ for $\lambda_{\mathrm{fcl}}$ (range $0.2990$--$0.3059$).
All variations remain below $2.5\%$ across one order of magnitude
of each coefficient, demonstrating that FDRMFL’s performance is
robust to the regularization weight settings and does not depend
on careful hyperparameter tuning.
In particular, the insensitivity of $\lambda_{\mathrm{kl}}$ across an order of magnitude indicates that the cross-modal alignment regularization is effective across a wide range of penalty strengths. The full sensitivity landscape is shown in Figure~\ref{fig:sensitivity}.

\paragraph{Reproducibility note.}
All experimental results are averaged over 10~random seeds with identical seeds applied to data splitting, weight initialization, and mini-batch sampling across all methods.
Tables~\ref{e2}--\ref{e1} decompose results by client to illustrate cross-client variance.

\subsection{Limitations and failure analysis}

FDRMFL achieves the lowest MSE in every target--client combination across both datasets, but the magnitude of improvement varies by target. Gains are largest when overlapping spectral bands create multimodal disambiguation challenges (e.g., corn oil and protein, $20$--$42\%$ improvement over the per-client best baseline) and comparatively smaller for well-conditioned targets whose smoother spectral--chemical relationships linear methods already approximate reasonably (e.g., corn moisture and starch, $8$--$30\%$). On very small datasets ($N{=}80$ for Corn), the limited per-client sample size increases stochastic variation, though FDRMFL retains consistent improvements across all targets. More broadly, FDRMFL is most beneficial when the data are genuinely multimodal with complementary cross-modal information, meaningful non-IID heterogeneity exists across clients, and the sample size is small relative to the feature dimensionality.

The current formulation also assumes full client participation in every communication round. If a client misses rounds, its local contrastive history buffer becomes stale; a natural mitigation is to reinitialize the buffer from the current global model upon rejoining, though we leave empirical validation of this strategy to future work.

We note that FDRMFL's privacy model is \emph{baseline federated privacy}: raw data never leave the local client, but we do not provide formal guarantees such as differential privacy or secure aggregation. Strengthening privacy to formal standards is an explicit direction for future work (Section~\ref{section4}).

\section{Conclusions}\label{section4}

We presented FDRMFL, a task-driven multimodal federated feature extraction
framework that jointly addresses task-relevant dimensionality reduction,
cross-modal alignment, and representation stability in non-IID federated
regression. The framework combines MSE prediction loss with three
complementary regularizers---a correlation-based MI surrogate, a symmetric
KL alignment penalty, and an InfoNCE-style contrastive anchor---in a
single unified local objective.

Experiments on three synthetic and two real-world NIR spectroscopy
datasets under non-IID federated partitions confirm the effectiveness
of the approach: FDRMFL reduces mean MSE by $33.8\%$ relative to the best
traditional baseline (PCA) and by $43.0\%$ relative to VAE,
with reduced cross-client variance.
In a separate comparison with five federated algorithms, FDRMFL
attains the lowest overall mean MSE with consistent performance
across all tasks.
Ablation and sensitivity analyses confirm that each component
contributes to performance and that results are robust to
hyperparameter settings.

Future work will pursue two directions:
(i)~incorporating differential privacy for formal privacy guarantees
beyond baseline data locality;
and (ii)~evaluating on more diverse multimodal data types under
stronger non-IID regimes.

\appendix
\section{Encoder architecture details}\label{app:encoders}

This appendix provides the detailed mathematical formulations of
the modality-specific encoders and prediction head referenced in
Section~\ref{sec:encoders}. All encoder outputs are projected to a
common dimension~$d$; parameters are included in the global model
$w^t$ and participate in federated aggregation.

\paragraph{Transformer encoder (text/sequential data).}
Input tokens are embedded and position-encoded to obtain
$X_{\text{text}}\!\in\!\mathbb{R}^{n\times d_{\text{model}}}$.
Scaled dot-product attention computes
\begin{equation}
\mathrm{Attention}(Q,K,V)
= \mathrm{Softmax}\!\!\left(\frac{QK^\top}{\sqrt{d_k}}\right)\!V
\end{equation}
where $Q\!=\!X_{\text{text}}W_Q$, $K\!=\!X_{\text{text}}W_K$,
$V\!=\!X_{\text{text}}W_V$ with
$W_Q,W_K,W_V\!\in\!\mathbb{R}^{d_{\text{model}}\times d_k}$.
Multi-head attention with $h$ heads concatenates individual heads
and projects via $W_O\!\in\!\mathbb{R}^{hd_k\times d_{\text{model}}}$:
\begin{equation}
\mathrm{MultiHead}(Q,K,V)
= \mathrm{Concat}(\mathrm{head}_1,\ldots,\mathrm{head}_h)\,W_O
\end{equation}
Each encoder layer additionally applies a position-wise feedforward
network $\mathrm{FFN}(x)=\max(0,xW_1+b_1)W_2+b_2$ and layer
normalization
$\mathrm{LN}(x)=\gamma\cdot(x-\mathbb{E}[x])/\!\sqrt{\mathrm{Var}(x)+\epsilon}+\beta$,
where $\gamma,\beta\!\in\!\mathbb{R}^{d_{\text{model}}}$ are
learnable.

\paragraph{CNN encoder (image data).}
Given an input tensor $X\!\in\!\mathbb{R}^{H\times W\times C_{\text{in}}}$,
the $k$-th output feature map of a convolutional layer is
\begin{equation}
Z_k = \sum_{c=1}^{C_{\text{in}}} X_c * K_{k,c} + b_k
\end{equation}
where $K_{k,c}\!\in\!\mathbb{R}^{k_h\times k_w}$ is the kernel weight
and $*$ denotes 2-D convolution:
$(X_c * K_{k,c})(i,j)=\sum_{p,q}X_c(i{+}p,j{+}q)\,K_{k,c}(p,q)$.
ReLU activation $a_k=\max(0,Z_k)$ and max-pooling
$\mathrm{MaxPool}(a_k)(i,j)=\max_{p,q\in\text{window}}a_k(p,q)$
follow each convolutional block. After flattening, a fully connected
layer maps the feature to dimension~$d$:
\begin{equation}
z_{\text{img}} = \max(0,\,X_{\text{flat}}\,W_{\text{fc}}+b_{\text{fc}})\,W_{\text{out}}+b_{\text{out}}
\end{equation}

\paragraph{LSTM encoder (time-series data).}
At each time step $t$, the LSTM computes forget, input, and output
gates from the concatenation $[h_{t-1},x_t]$:
\begin{align}
f_t &= \sigma(W_f[h_{t-1},x_t]+b_f) \\
i_t &= \sigma(W_i[h_{t-1},x_t]+b_i) \\
\tilde{C}_t &= \tanh(W_C[h_{t-1},x_t]+b_C) \\
C_t &= f_t\odot C_{t-1}+i_t\odot\tilde{C}_t \\
o_t &= \sigma(W_o[h_{t-1},x_t]+b_o) \\
h_t &= o_t\odot\tanh(C_t)
\end{align}
where $\sigma$ is the sigmoid function, $\odot$ denotes element-wise
multiplication, and all gate weight matrices
$W_{\{f,i,C,o\}}\!\in\!\mathbb{R}^{d_h\times(d_h+d_{\text{in}})}$.
A bidirectional variant is used in practice; the final hidden state
is projected to dimension~$d$ via
$z_{\text{seq}}=h_T W_{\text{lstm-out}}+b_{\text{lstm-out}}$.

\paragraph{MLP encoder (vector data).}
For vector-valued input $X_{\text{vec}}\!\in\!\mathbb{R}^{d_{\text{in}}}$,
an $L$-layer MLP applies alternating linear transforms and ReLU
activations:
\begin{equation}
a_l = \max(0,\,W_l\,a_{l-1}+b_l),\quad l=1,\ldots,L
\end{equation}
with $a_0=X_{\text{vec}}$, $W_l\!\in\!\mathbb{R}^{d_l\times d_{l-1}}$,
and $b_l\!\in\!\mathbb{R}^{d_l}$. The output layer maps
$a_L$ to dimension~$d$:
$z_{\text{vec}}=W_{L+1}\,a_L+b_{L+1}$.

\paragraph{Prediction head $f$.}
The prediction head is a four-layer MLP that maps the fused
representation $Z_i\!\in\!\mathbb{R}^d$ to a scalar regression
output:
\begin{equation}
\hat{y}_i = W_{f,L_f+1}\,a_{f,L_f}+b_{f,L_f+1},
\quad a_{f,l}=\max(0,\,W_{f,l}\,a_{f,l-1}+b_{f,l})
\end{equation}
where $W_{f,L_f+1}\!\in\!\mathbb{R}^{1\times d_{f,L_f}}$ produces
the scalar output. All prediction-head parameters participate in
federated aggregation.

\bibliographystyle{cas-model2-names}
\bibliography{name}

\end{document}